\documentclass[
]{ceurart}

\usepackage{listings}
\begin{document}

\copyrightyear{2025}
\copyrightclause{Copyright for this paper by its authors.
  Use permitted under Creative Commons License Attribution 4.0
  International (CC BY 4.0).}

\conference{International Workshop on AI Governance: Alignment, Morality and Law (AIGOV) 2025}

\title{MINT-Demo: Membership Inference Test Demonstrator}

\author[1]{Daniel DeAlcala}[%
email=daniel.dealcala@uam.es
]
\author[1]{Aythami Morales}[%
email=aythami.morales@uam.es,
]
\author[1]{Julian Fierrez}[%
email=julian.fierrez@uam.es,
]
\author[1]{Gonzalo Mancera}[%
email=gonzalo.mancera@uam.es,
]
\author[1]{Ruben Tolosana}[%
email=ruben.tolosana@uam.es,
]
\author[1]{Ruben Vera-Rodriguez}[%
email=ruben.vera@uam.es,
]
\address[1]{Biometrics and Data Pattern Analytics Lab, Universidad Autonoma de Madrid, 28049 Madrid, Spain}

\begin{abstract}
We present the Membership Inference Test Demonstrator, to emphasize the need for more transparent machine learning training processes. MINT is a technique for experimentally determining whether certain data has been used during the training of machine learning models. We conduct experiments with popular face recognition models and $5$ public databases containing over $22$M images. Promising results, up to 89\% accuracy are achieved, suggesting that it is possible to recognize if an AI model has been trained with specific data. Finally, we present a MINT platform as demonstrator of this technology aimed to promote transparency in AI training\footnote{\href{https://ai-mintest.org/}{https://ai-mintest.org/}}.
\end{abstract}


\maketitle


\section{Introduction}

The unauthorized use of personal or copyrighted material to train AI models may infringe upon the rights of individuals. Moreover, the generated output of AI models trained on this data may blur the line between original and derived works, raising concerns of plagiarism and copyright infringement.

On June 2024, the European Union has finally published the highly anticipated Artificial Intelligence Act \cite{AIact} requiring AI providers to ensure robust protection of fundamental citizen rights. The regulation mandates the registration of AI models in an EU database and grants national authorities the power to request access to trained models and their training data. This regulation enforces transparency and calls for new auditing tools to ensure secure AI deployment in Europe.
 
%This new regulation is a game-changer imposing transparency as a must to deploy AI technologies in Europe. The AI act urges to develop new auditing tools to monitor AI technologies and their secure deployment in our society.

These considerations lead us to the main objective of this work, which is to propose a platform to detect the data used to train AI models. Currently, developers can hide behind the weights of their network to bypass regulations and conceal the use of training data from users.This approach seeks to unveil AI training processes, ensuring alignment with legislation and citizen rights.
% Using the concepts outlined here, we aim to unveil AI training processes and align such training with current legislation and citizen rights.

\begin{figure}
\centering
\includegraphics[width=0.50\linewidth]{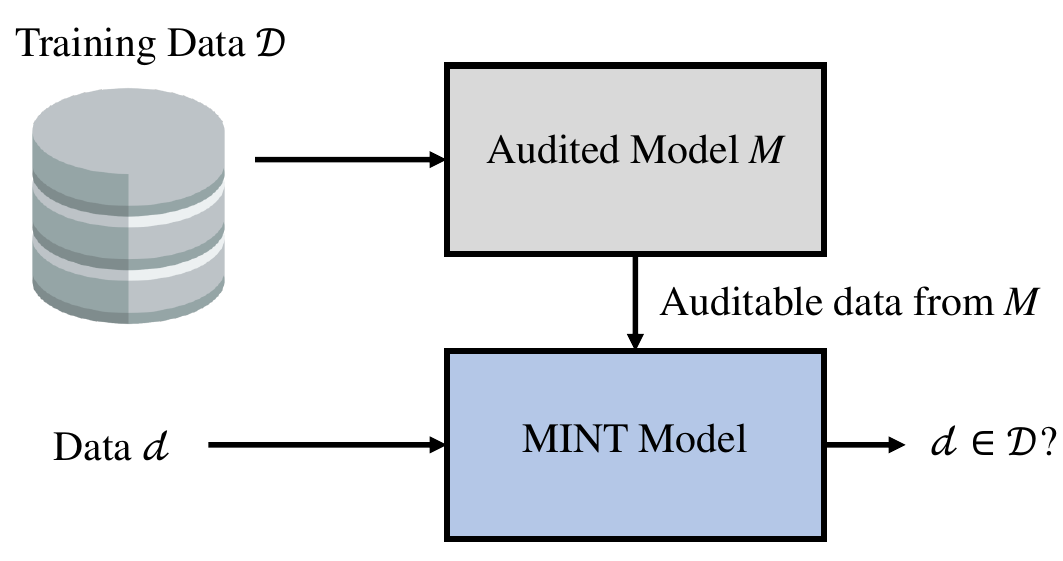} % Reduce the figure size so that it is slightly narrower than the column. Don't use precise values for figure width.This setup will avoid overfull boxes.
\caption{Block diagram of the Membership Inference Test.} %Model ($T$) is trained to predict if an specific data ($d$) was used during the training process of a Audited Artificial Intelligence Model ($M$) trained with a database ($D$). The input of the MINT Model is auxiliary data (e.g., activations maps for data samples $d$) obtained from $M$.}
\label{Block_diagram_full}
\end{figure}

The main contributions can be summarized as follows:

\begin{itemize}
    \item Introduced MINT, a method to detect data usage during AI training (Fig. \ref{Block_diagram_full}), aiding compliance with AI legislation and protecting citizen rights.
    \item Conducted experiments on $5$ public datasets with $22$M+ images, achieving up to $89\%$ accuracy in Membership Inference Tests, highlighting its challenges and benefits.
    \item Developed an interactive web platform with real models to promote AI transparency.  
\end{itemize}

\section{Membership Inference Test}

Let us consider a training dataset $\mathcal{D}$, an external dataset $\mathcal{E}$ and a collection of samples $d \in \mathcal{D} \cup \mathcal{E}$. We assume a learned model $M$ that is trained for a specific task (text generation, face recognition, etc.) using the dataset $\mathcal{D}$. For any input data $d$, the model $M$ generates an outcome $y$ based on $d$ and a set of parameters $\textbf{w}$ ($y=M(d|\textbf{w})$). From that model $M$ we assume access to partial information that we call Auxiliary Auditable Data ${AAD}$ (e.g., activation maps of specific layers in a Neural Network), which is based on $d$ and a subset of parameters $\textbf{w}' \in \textbf{w} $ (${AAD}=N(d|\textbf{w}')$). 

\begin{figure}[h]
\centering
\includegraphics[width=\linewidth]{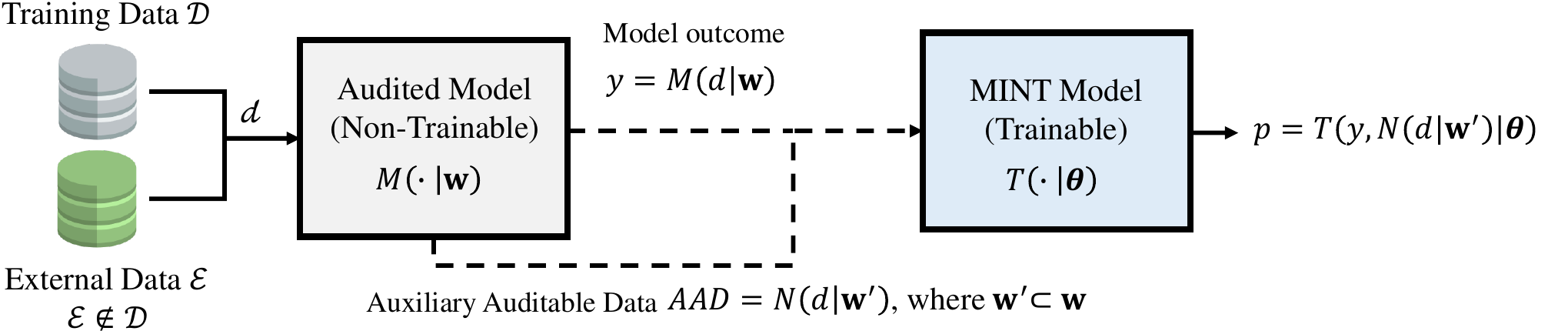} % Reduce the figure size so that it is slightly narrower than the column. Don't use precise values for figure width.This setup will avoid overfull boxes.
\caption{The MINT Model ($T$) predicts whether specific data ($d$) was used to train an Audited AI Model ($M$), using Auxiliary Auditable Data (e.g., activation maps) and/or the model outcome from $M$.}
\label{Block_Diagram_lite}
\end{figure}

The Membership Inference Test (MINT) aims to determine if data $d$ was used to train the model $M$, i.e., if $d$ belongs to $\mathcal{D}$ or $\mathcal{E}$ ($\mathcal{E} \notin \mathcal{D}$). To this end, an authorized authority employs the available information to train an auditing model ($T(\cdot|\theta)$) as indicated in the workflow in Fig. \ref{Block_Diagram_lite}.

% \begin{itemize}
%     \item Audited Model ($M$): a learned model defined by an architecture and a set of parameters $\textbf{w}$.
%     \item Training Data ($\mathcal{D}$): collection of data used to train $M$.
%     \item External Data ($\mathcal{E}$): any data out of the collection ($\mathcal{D}$).
%     \item Model Outcome ($y = M(d|\textbf{w})$): final outcome of $M$ that results from processing an input data $d$.
%     \item Auxiliary Auditable Data (${AAD} = N(d|\textbf{w}')$): intermediate outcomes of $M$ that results from processing an input data $d$ using a subset $\textbf{w}'$ of the parameters $\textbf{w}$. %The model outcome can be seen as the case where $N(d|\textbf{w}')=M(d|\textbf{w})$. 
%     \item MINT Model ($T$): a model defined by an architecture and a set of parameters $\theta$ trained using Auxiliary Auditable Data ($N(d|\textbf{w}')$) and/or Model Outcomes ($M(d|\textbf{w})$) of $M$ obtained from the two subsets of samples $\mathcal{D}$ and $\mathcal{E}$.
% \end{itemize}

\section{Experiments}

Here we report experiments with a popular face recognition model from the InsigthFace project\footnote{\href{https://insightface.ai/}{https://insightface.ai/}}. Similar experiments have been conducted with other face recognition models and can be tested on the website\footnote{\href{https://ai-mintest.org/}{https://ai-mintest.org/}}. The face recognition model ($M$ in Fig. \ref{Block_diagram_full}) is a ResNet-100 network \cite{han2017deep}, trained on the Glint360k database \cite{an2021partial} with CosFace loss function \cite{wang2018cosface}. This database comprise 17M images ($\mathcal{D}$ in Fig. \ref{Block_diagram_full}). 

We propose two different MINT model architectures:
\begin{enumerate}
    \item \textbf{Vanilla MINT Model}: An MLP consisting of three fully connected layers—input-size neurons (varying by Auxiliary Auditable Data), 64 neurons, and 1 neuron. A dropout layer (0.3 rate) and an L1 regularizer (0.1) are applied between layers.
    \item \textbf{CNN MINT Model}: A CNN with two convolutional layers (64 and 128 filters) followed by two fully connected layers sized to the convolution output.
\end{enumerate}

We included the IJB-C \cite{IJB}, FDDB \cite{jain2010fddb}, GANDiffFace \cite{deandres2024frcsyn}, and Adience \cite{Adience} databases as external Data ($\mathcal{E}$) to train and test the MINT model $T$. For the Auxiliary Auditable Data, we used activations from various layers in $M$. In the Vanilla MINT Model, we extract the maximum value from each activation map at different depths, forming a vector whose size depends on the number of filters in the selected layer. The CNN MINT Model, however, analyzes activations directly, using the full activation maps without vectorizing them.

Table \ref{Table:AccForLayerNN} presents the classification accuracy for the Vanilla MINT model. The columns represent the number of samples used to train the MINT model ($T$), and the rows show the depth of the selected activation maps (Auxiliary Auditable Data). We focus on the final convolutional layer of each of the 4 major ResNet-100 blocks (first to fourth layers). The table also includes the ``Model Outcome'' (model's output embedding) and ``All Conv Layers'' (concatenated Conv Layers). The classification accuracy varies depending on the available Auxiliary Auditable Data and amount of data, with "all layers" yielding the best performance (up to 84\%). The best individual results come from layers closest to the input and output, while intermediate layers show poorer results.

\begin{table}[t]
\centering
\begin{tabular}{@{}lccc@{}}
\hline
Auditable Data & \multicolumn{1}{l}{$1$K samples} & $50$K samples & $100$K samples \\ \hline

Conv Layer \#1 & 0.62 & 0.80 & 0.80 \\
Conv Layer \#2 & 0.56 & 0.67 & 0.68 \\
Conv Layer \#3 & 0.56 & 0.58 & 0.59 \\
Conv Layer \#4 & 0.73 & 0.76 & 0.76 \\ %\midrule
Model Outcome & 0.67 & 0.78 & 0.78 \\
\textbf{All Conv Layers} & \textbf{0.76} & \textbf{0.82} & \textbf{0.84} \\ \hline
\end{tabular}
\caption{Classification accuracy using the Vanilla MINT Model. The MINT model was trained with a variable number of samples ranging from $1K$ to $100k$.}
\label{Table:AccForLayerNN}
\end{table}

\begin{table}[t]
\centering
\begin{tabular}{@{}lccc@{}}
\hline
Auditable Data & \multicolumn{1}{l}{$1$K samples} & $50$K samples & $100$K samples \\ \hline

\textbf{Conv Layer \#1} & \textbf{0.88} & \textbf{0.89} & \textbf{0.89} \\
Conv Layer \#2 & 0.85 & 0.86 & 0.86 \\
Conv Layer \#3 & 0.68 & 0.71 & 0.75 \\
Conv Layer \#4 & 0.68 & 0.70 & 0.74 \\ \hline
\end{tabular}
\caption{Classification accuracy using the CNN MINT Model. The MINT model was trained with a variable number of samples ranging from $1K$ to $100k$. }
\label{Table:AccForLayerCNN}
\end{table}

Table \ref{Table:AccForLayerCNN} presents the results for the CNN MINT Model. Notably, there are no results for the Model Outcome, as CNN architectures cannot be applied directly to the output vector. Similarly, the row for concatenating convolutional layers is missing due to the varying resolutions of activation maps, making concatenation impractical—unlike the Vanilla MINT Model where vectorization made it feasible. In this architecture, the best performance is achieved with the layer closest to the input, with accuracy decreasing towards the output. The CNN MINT Model achieves 89\% accuracy, outperforming the Vanilla model's 84\%.

\section{Demonstrator and Future}

Our MINT demonstrator can be accessed via web here: https://ai-mintest.org/. There you can upload images and receive reports on the likelihood that these images were used to  train multiple AI models. This demonstrator now includes a limited initial set of popular image models, which we expect to grow across other modalities including image \cite{dealcala2024my} and text \cite{mancera2025my}. Future work will investigate other learning architectures for MINT and study the factors impacting membership inference \cite{dealcala2024comprehensive} in general and for relevant AI models.

\section{Acknowledgement}

This work has been supported by projects BBforTAI (PID2021-127641OB-I00 MICINN/FEDER), Cátedra ENIA UAM-VERIDAS (NextGenerationEU PRTR TSI-100927-2023-2), and Comunidad de Madrid (ELLIS Unit Madrid). D. DeAlcala is supported by a FPU Fellowship (FPU21/05785) from the Spanish MIU. G. Mancera is supported by FPI-PRE2022-104499 MICINN/FEDER.

\bibliography{main}

\end{document}